\documentclass{elsarticle}
\usepackage{textcomp}
\usepackage{etoolbox}
\makeatletter
\patchcmd{\ps@pprintTitle}{\footnotesize\itshape
       Preprint submitted to \ifx\@journal\@empty Elsevier
       \else\@journal\fi\hfill\today}{\footnotesize\itshape
       Accepted for Medical Image Analysis (Special Issue)\hfill\today}{}{}
\makeatother

\usepackage{hyperref}

\journal{Medical Image Analysis}

\usepackage[utf8]{inputenc} 
\usepackage[T1]{fontenc}    
\usepackage{hyperref}       
\usepackage{url}            
\usepackage{booktabs}       
\usepackage{amsmath}
\usepackage{amsfonts}       
\usepackage{bm}
\usepackage{soul}

\usepackage{nicefrac}       
\usepackage{microtype}      
\usepackage{comment}
\usepackage{graphicx}
\graphicspath{figures/}
\usepackage{xcolor}
\hypersetup{hidelinks}
\usepackage{colortbl,array} 
\usepackage{multirow,bigdelim}
\usepackage{booktabs}
\usepackage{caption}
\usepackage{lipsum}  
\usepackage{todonotes}
\usepackage{subcaption}
\usepackage{float}

\newcommand{\bluereview}[1]{{\color{black}{#1}}}
\newcommand{\blue}[1]{{\color{black}{#1}}}
\newcommand{\redreview}[1]{\phantom{#1}}

\newcommand{\R}[0]{\mathbb{R}} 

\providecommand{\mb}[1]{\mathbf{#1}}

\usepackage{numcompress}\biboptions{authoryear}

\begin{document}

\begin{frontmatter}

\title{Attention Gated Networks:\\ Learning to Leverage Salient Regions in Medical Images}

\author[mymainaddress]{Jo Schlemper\fnref{myfootnote}} 

\author[mymainaddress,mysecondaryaddress]{Ozan Oktay\fnref{myfootnote}} 
\author[mysecondaryaddress]{Michiel Schaap}
\author[mythirdaddress]{Mattias Heinrich}
\author[mymainaddress]{Bernhard Kainz}
\author[mymainaddress]{Ben Glocker} 
\author[mymainaddress]{Daniel Rueckert}

\fntext[myfootnote]{The corresponding authors contributed equally.}
\fntext[]{\textcopyright 2019. This manuscript version is made available under the CC-BY-NC-ND 4.0 license (\url{http://creativecommons.org/licenses/by-nc-nd/4.0/})}
\address[mymainaddress]{BioMedIA, Imperial College London, SW7 2AZ, London, UK}
\address[mysecondaryaddress]{HeartFlow, Redwood City, CA 94063, USA}
\address[mythirdaddress]{Medical Informatics, University of Luebeck, DE}

\begin{abstract}
We propose a novel attention gate (AG) model for medical image analysis that automatically learns to focus on target structures of varying shapes and sizes. Models trained with AGs implicitly learn to suppress irrelevant regions in an input image while highlighting salient features useful for a specific task. This enables us to eliminate the necessity of using explicit external tissue/organ localisation modules when using convolutional neural networks (CNNs). AGs can be easily integrated into standard CNN models such as VGG  or U-Net architectures with minimal computational overhead while increasing the model sensitivity and prediction accuracy. The proposed AG models are evaluated on a variety of tasks, including medical image classification and segmentation. For classification, we demonstrate the use case of AGs in scan plane detection for fetal ultrasound screening. We show that the proposed attention mechanism can provide efficient object localisation while improving the overall prediction performance by reducing false positives. For segmentation, the proposed architecture is evaluated on two large 3D CT abdominal datasets with manual annotations for multiple organs. Experimental results show that AG models consistently improve the prediction performance of the base architectures across different datasets and training sizes while preserving computational efficiency. Moreover, AGs guide the model activations to be focused around salient regions, which provides better insights into how model predictions are made. The source code for the proposed AG models is publicly available.
\end{abstract}


\end{frontmatter}


\section{Introduction}
Automated medical image analysis has been extensively studied in the medical imaging  community due to the fact that manual labelling of large amounts of medical images is a tedious and error-prone task. Accurate and reliable solutions are required to increase clinical work flow efficiency and support decision making through fast and automatic extraction of quantitative measurements.

With the advent of convolutional neural networks (CNNs), near-radiologist level performance can be achieved in automated medical image analysis tasks including classification of Alzheimer's disease \citep{Sarraf070441}, skin lesions \citep{esteva2017dermatologist,kawahara2016multi} and echo-cardiogram views \citep{madani2018fast}, lung nodule detection in CT/X-ray \citep{liao2017evaluate,DBLP:journals/corr/abs-1801-09555} and cardiac MR segmentation \citep{bai2017human}. An extensive list of applications can be found in \citep{DBLP:journals/corr/LitjensKBSCGLGS17,zaharchuk2018deep}. High representation power, fast inference, and weight sharing properties have made CNNs the de facto standard for image classification and segmentation. 

Methods for existing applications rely heavily on multi-stage, cascaded CNNs when the target organs show large inter-patient variation in terms of shape and size. Cascaded frameworks extract a region of interest (ROI) and make dense predictions on that particular ROI. The application areas include cardiac MRI \citep{khened2018fully}, cardiac CT \citep{payer2017multi}, abdominal CT \citep{roth2017hierarchical,roth2018media} segmentation, and lung CT nodule detection \citep{liao2017evaluate}. However, this approach leads to excessive and redundant use of computational resources and model parameters; for instance, similar low-level features are repeatedly extracted by all models within the cascade. 

To address this general problem, we propose a simple and yet effective solution, named \emph{attention gates} (AGs). \bluereview{By incorporating AGs into standard CNN models, model parameters and intermediate feature maps are expected to be utilised more efficiently while minimising the necessity of cascaded models to solve localisation and classification tasks separately. In more detail, AGs automatically learn to focus on target structures without additional supervision. At test time, these gates generate soft region proposals implicitly on-the-fly and highlight salient features useful for a specific task. In return, the proposed AGs improve model sensitivity and accuracy for global and dense label predictions by suppressing feature activations in irrelevant regions. In this way, the necessity of using an external organ localisation module can be eliminated while maintaining the high prediction accuracy. In addition, they do not introduce significant computational overhead and do not require a large number of model parameters as in the case of multi-model frameworks.} CNN models with AGs can be trained from scratch in a standard way similar to the training of fully convolutional network (FCN) models. Similar attention mechanisms have been proposed for natural image classification \citep{jetley2018learn} and captioning \citep{anderson2017bottom} to perform adaptive feature pooling, where model predictions are conditioned only on a subset of selected image regions. In this paper, we generalise this design and propose image-grid based gating that allows attention coefficients to be specific to local regions. 

We demonstrate the performance of AG in real-time fetal ultrasound scan plane detection and CT pancreas segmentation. The first task is challenging due to low interpretability of the images and localising the object of interest is key to successful classification of the plane. To this end, we incorporate AGs into a variant of a VGG network, termed AG-Sononet, to demonstrate that attention mechanism can automatically localise the object of interest and improve the overall classification performance. The second task of pancreas segmentation is challenging due to low tissue contrast and large variability in organ shape and size. Moreover, we extend a standard U-Net architecture (\emph{Attention U-Net}). We choose to evaluate our implementation on two commonly used benchmarks: TCIA Pancreas $CT$-$82$ \citep{tciapancreas} and multi-class abdominal $CT$-$150$. The results show that AGs consistently improve prediction accuracy across different datasets and training sizes while achieving state-of-the-art performance without requiring multiple CNN models. 
%
%

\vspace{-1.0 mm}
\subsection{Related Work}
\blue{\textbf{Attention Gates:}} AGs are commonly used in classification tasks such as in the analysis of \blue{citation graphs \citep{velivckovic2017graph}} and natural images \citep{jetley2018learn, wang2017residual}. Similarly in the context of natural language processing (NLP), such as image captioning \citep{anderson2017bottom} and \blue{machine translation \citep{bahdanau2014neural, luong2015effective, shen2017disan, vaswani2017attention}}, there have been several use cases of soft-attention models to efficiently use the given context information. In particular, given a sequence of text and a current word, a task is to extract a next word in a sentence generation or translation. The idea of attention mechanisms is to generate a \emph{context} vector which assigns weights on the input sequence. Thus, the signal highlights the salient feature of the sequence conditioned on the current word while suppressing the irrelevant counter-parts, making the prediction more contextualised.

Initial work on attention modelling has explored salient image regions by interpreting gradient of output class scores with respect to the input image. Trainable attention, on the other hand, is enforced by design and categorised as hard- and soft-attention. Hard attention \citep{mnih2014recurrent}, e.g. iterative region proposal and cropping, is often non-differentiable and relies on reinforcement learning for parameter updates, which makes model training more difficult. \citet{ypsilantis2017learning} used recursive hard-attention to detect anomalies in chest X-ray scans. Contrarily, soft attention is probabilistic, end-to-end differentiable, and utilises standard back-propagation without need for posterior sampling. For instance, additive soft attention is used in sentence-to-sentence translation \citep{bahdanau2014neural, shen2017disan} and more recently applied to image classification \citep{jetley2018learn, wang2017residual}. 

In computer vision, attention mechanisms are applied to a variety of problems, including image classification \citep{jetley2018learn, wang2017residual, zhao2017survey}, segmentation \citep{DBLP:journals/corr/RenZ16}, action recognition \citep{liu2017global,DBLP:journals/corr/PeiBTM16,wang2017non}, image captioning \citep{lu2017knowing,xu2015show}, and visual question answering \citep{DBLP:journals/corr/NamHK16,DBLP:journals/corr/YangHGDS15}. \citet{hu2017squeeze} used channel-wise attention to highlight important feature dimensions, which was the top-performer in the ILSVRC 2017 image classification challenge. Similarly, non-local self attention was used by \citet{wang2017non} to capture long range dependencies. 

In the context of medical image analysis, attention models have been exploited for medical report generation \citep{zhang2017tandemnet,zhang2017mdnet} as well as joint image and text classification \citep{DBLP:journals/corr/abs-1801-04334}. However, for standard medical image classification, despite often the information to be classified are extremely localised, only a handful of works use attention mechanisms  \citep{guan2018diagnose,pesce2017learning}. In these methods, either bounding box labels are available to guide the attention, or local context is extracted by a hard-attention model (i.e. region proposal followed by hard-cropping). 

\textbf{2D Ultrasound Scan Plane Detection:} Fetal ultrasound screening is an important diagnostic protocol to detect abnormal fetal development. During screening examination, multiple anatomically standardised~\citep{FASP} scan planes are used to obtain biometric measurements as well as identifying abnormalities such as lesions. Ultrasound suffers from low signal-to-noise ratio and image artefacts. As such, diagnostic accuracy and reproducibility is limited and requires a high level of expert knowledge and training. In the past, several approaches were proposed \citep{chen2015automatic,yaqub2015guided}, however, they are computationally expensive and cannot be deployed for the real-time application. More recently, \citet{baumgartner2016real} proposed a CNN architecture called \emph{Sononet}. It achieves very good performance in real-time plane detection, retrospective frame retrieval (retrieving the most relevant frame) and weakly supervised object localisation. However, it suffers from low recall value in differentiating different planar views of the cardiac chambers, which requires the method to be able to exploit the subtle differences in the local structure and it makes the problem challenging.

\textbf{Pancreas Segmentation in 3D-CT Images:} Early work on pancreas segmentation from abdominal CT used statistical shape models \citep{cerrolaza2016soft, saito2016joint} or multi-atlas techniques \citep{oda20173d, wolz2013automated}. In particular, atlas approaches benefit from implicit shape constraints enforced by propagation of manual annotations. However, in public benchmarks such as the TCIA dataset \citep{tciapancreas}, Dice similarity coefficients (DSC) for atlas-based frameworks are relatively low, ranging from $69.6\%$ to $73.9\%$ \citep{oda20173d, wolz2013automated}. A classification based framework was proposed by \citet{zografos2015hierarchical} to remove the dependency of atlas to image registration. Recently, cascaded multi-stage CNN models \citep{roth2017hierarchical,roth2018media,zhou2017fixed} have been proposed to address the problem. Here, an initial coarse-level model (e.g. U-Net or Regression Forest) is used to obtain a ROI and then a cropped ROI is used for segmentation refinement by a second model. Similarly, combinations of 2D-FCN and recurrent neural network (RNN) models are utilised by \citet{cai2017improving} to exploit dependencies between adjacent axial slices. These approaches achieve state-of-the-art performance in the TCIA benchmark ($81.2\% - 82.4\%$ DSC). \bluereview{Without using a cascaded framework, the performance drops between $2.0$ and $4.4$ DSC points}. Recently, \citet{yu2017saliency} proposed an iterative two-stage model that recursively updates local and global predictions, and both models are trained end-to-end. Besides standard FCNs, dense connections \citep{gibson2017towards} and sparse convolutions \citep{heinrich2018ternarynet, heinrich2017briefnet} have been applied to the CT pancreas segmentation problem. Dense connections and sparse kernels reduce computational complexity by requiring less number of non-zero parameters. 

\subsection{Contributions} In this paper, we propose a novel soft-attention gating module that can be utilised in CNN based standard image analysis models for dense label predictions. Additionally, we explore the benefit of AGs to medical image analysis, in particular, in the context of image classification and segmentation. The contributions of this work can be summarised as follows:

\begin{itemize}
	
	\item We take the attention approach proposed by \citet{jetley2018learn} a step further by proposing grid-based gating that allows attention gates to be more specific to local regions. This improves performance compared to gating based on a global feature vector. Moreover, our approach is not only limited to adaptive pooling \citep{jetley2018learn} but can be also used for dense predictions as in segmentation networks.  
	
	\item We propose one of the first use cases of soft-attention in a feed-forward CNN model applied to a medical imaging task that is end-to-end trainable. The proposed attention gates can replace hard-attention approaches used in image classification \citep{ypsilantis2017learning} and external organ localisation models in image segmentation frameworks \citep{khened2018fully, oda20173d,roth2017hierarchical,roth2018media}. This also eliminates the need for any bounding box labels and backpropagation-based saliency map generation used by \citet{baumgartner2016real}.

	\item For classification, we apply the proposed model to real-time fetal ultrasound scan plane detection and show its superior classification performance over the baseline approach. We show that attention maps can used for fast (weakly-supervised) object localisation, demonstrating that the attended features indeed correlate with the anatomy of interest.

	\item For segmentation, an extension to the standard U-Net model is proposed that provides increased sensitivity without the need of complicated heuristics, while not sacrificing specificity. We demonstrate that accuracy improvements when using U-Net are consistent across different imaging datasets and training sizes. 

	\item We demonstrate that the proposed attention mechanism provides fine-scale attention maps that can be visualised, with minimal computational overhead, which helps with interpretability of predictions. 

\end{itemize}

\section{Methodology}


\subsection{Convolutional Neural Network} 

CNNs are now the state-of-the-art method for many tasks including classification , localisation and segmentation \citep{bai2017human,kamnitsas2017efficient,kamnitsas2018ensembles,lee2015deeply,DBLP:journals/corr/LitjensKBSCGLGS17,long2015fully,ronneberger2015u,roth2017hierarchical,roth2018media,xie2015holistically,zaharchuk2018deep}. CNNs outperform traditional approaches in medical image analysis while being an order of magnitude faster than, e.g., graph-cut and multi-atlas segmentation techniques \citep{wolz2013automated}. The success of CNNs is attributed to the fact that (I) domain specific image features are learnt using stochastic gradient descent (SGD) optimisation, (II) learnt kernels are shared across all pixels, and (III) image convolution operations exploit the structural information in medical images in an optimal fashion. However, it remains difficult to reduce false-positive predictions for small objects that show large shape variability. In such cases, in order to improve the accuracy, current frameworks \citep{guan2018diagnose,khened2018fully,roth2017hierarchical,roth2018media} rely on additional preceding object localisation models to simplify the task into separate localisation and subsequent classification/segmentation steps, or guide the localisation using weak labels \citep{pesce2017learning}. Here, we demonstrate that the same objective can be achieved by integrating attention gates (AGs) in a standard CNN model. This does not require the training of multiple models and a large number of extra model parameters. In contrast to the localisation model in multi-stage CNNs, AGs progressively suppress feature responses in irrelevant background regions without the requirement to crop a ROI between networks.

\begin{figure}[t]
	\centering
	\includegraphics[width=0.95\textwidth]{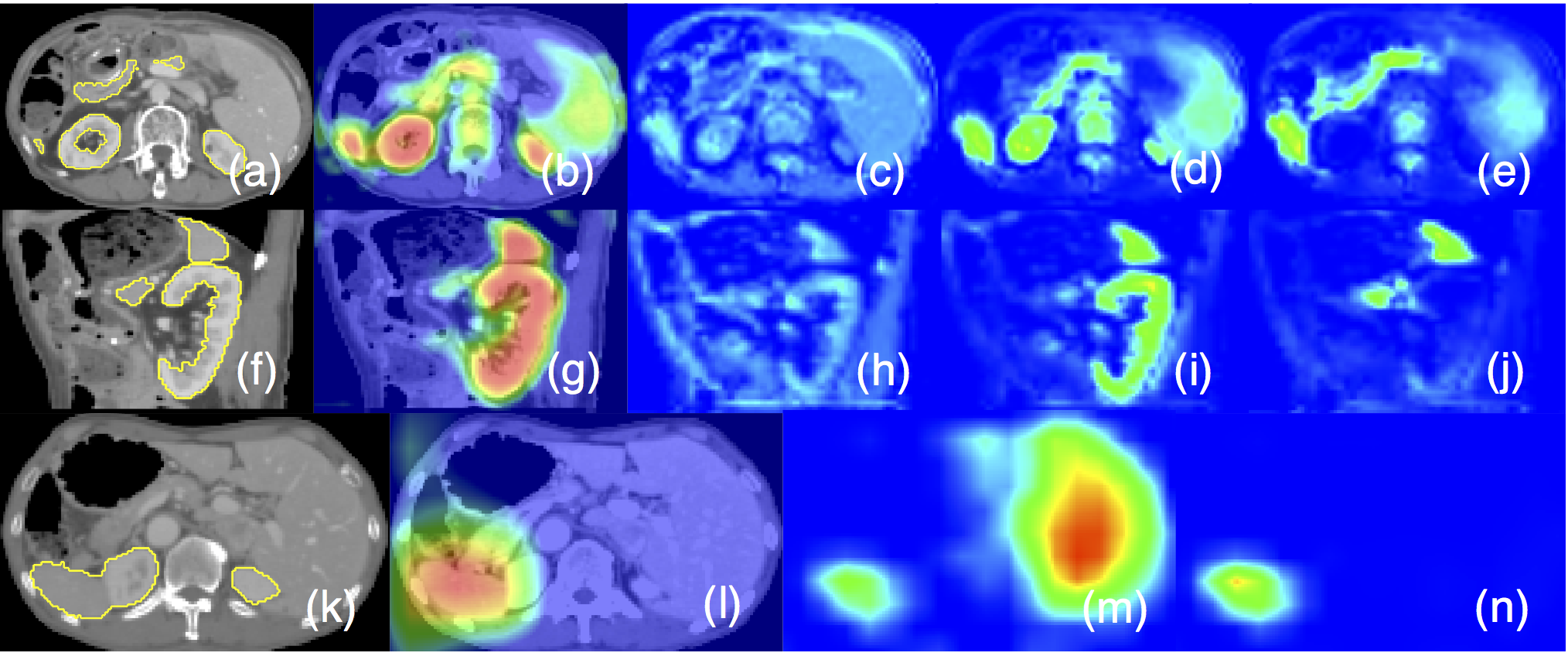}
	\caption{\linespread{1.2}\small Axial (a) and sagittal (f) views of a 3DCT scan, (b,g) attention coefficients, \bluereview{image feature activations before (c,h) and after attention gating (d,e,i,j)}. Similarly, (k-n) visualise the gating on a coarse scale skip connection. The filtered feature activations (d,e,i,j) are collected from multiple AGs, where a subset of organs is selected by each gate and activations consistently correspond to specific structures across different scans.}
	\label{fig:attention_activations}
\end{figure}

\subsection{Attention Gate Module} 

We now introduce \emph{Attention Gate} (AG), which is a mechanism which can be incorporated in any existing CNN architecture. Let $\mb{x}^l = \{ \mb{x}_i^{l} \}_{i=1}^n$ be the activation map of a chosen layer $l\in \{1, \dots, L\}$, where each $\mb{x}_i^{l}$ represents the pixel-wise feature vector of length $F_l$ (i.e. the number of channels). For each $\mb{x}_i^{l}$, AG computes coefficients $\alpha^l = \{\alpha_i^l \}_{=1}^n$, where $\alpha_i^l \in [0, 1]$, in order to identify salient image regions and prune feature responses to preserve only the activations relevant to the specific task as shown in Figure \ref{fig:attention_activations}. The output of AG is $\hat{\mb{x}}^l =\{ \alpha_i^l \mb{x}_i^l \}_{i=1}^n$, where each feature vector is scaled by the corresponding attention coefficient. 

The attention coefficients $\alpha_i^l$ are computed as follows: In standard CNN architectures, to capture a sufficiently large receptive field and thus, semantic contextual information, the feature-map is gradually downsampled. The features on the coarse spatial grid level identify location of the target objects and model their relationship at global scale. \bluereview{Let $\mb{g} \in \mathbb{R}^{F_g}$ be such global feature vector and provide information to AGs to disambiguate task-irrelevant feature content in $\mb{x}_i^{l}$.} The idea is to consider each $\mb{x}_i^{l}$ and $\mb{g}$ jointly to attend the features at each scale $l$ that are most relevant to the objective being minimised. 

\begin{figure}[!t]
	\centering
	\includegraphics[width=1.0\textwidth]{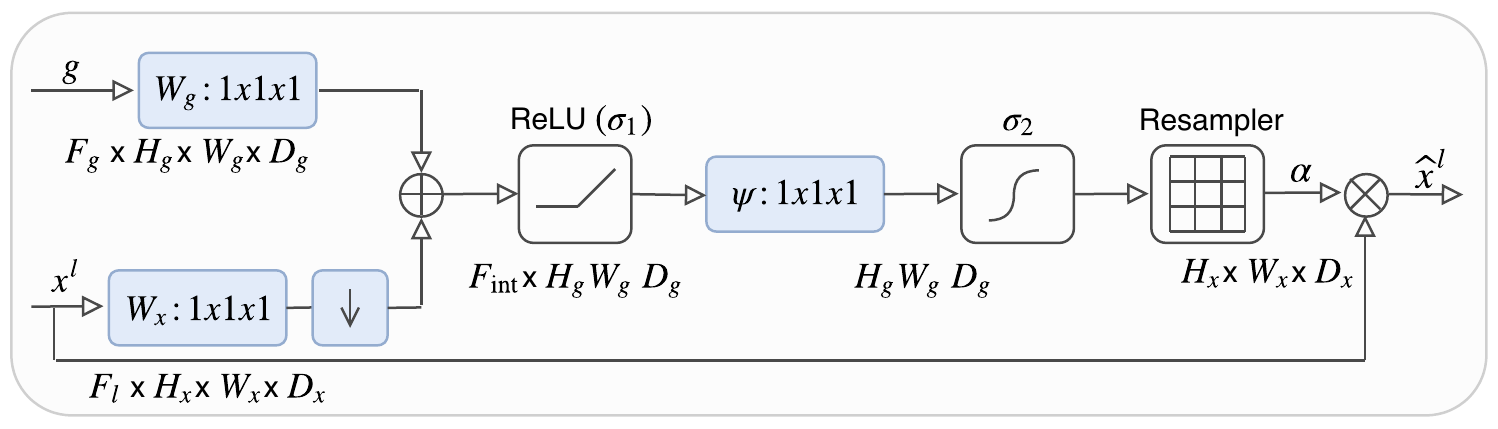}
	\caption{Schematic of the proposed additive attention gate (AG). Input features ($x^l$) are scaled with attention coefficients ($\alpha$) computed in AG. Spatial regions are selected by analysing both the activations and contextual information provided by the gating signal ($\,g\,$) which is collected from a coarser scale. Grid resampling of attention coefficients is performed using trilinear interpolation.}
	\label{fig:attentionblock}
\end{figure}

\bluereview{There are two commonly used attention types: multiplicative \citep{luong2015effective} and additive attention \citep{bahdanau2014neural}. The former is faster to compute and more memory-efficient in practice since it can be implemented as a matrix multiplication. However, additive attention is experimentally shown to be performing better for large dimensional input features \citep{britz2017massive}. For this reason, we use the latter to obtain the gating coefficient as can be seen in Figure \ref{fig:attentionblock}, which is formulated as follows:}
\begin{gather}
q^l_{att,i} = \bm{\psi}^T \left(\, \sigma_1\,(\,\bm{W}_x^T \bm{x}_i^l + \bm{W}_g^T \bm{g} + \bm{b}_{xg}\,)\,\right) + b_{\psi} \\
\alpha^l = \sigma_2 (\, q^l_{att}(\bm{x}^l\,,\, \bm{g} \,;\, \bm{\Theta}_{att}) \,), 
\end{gather}
where $\sigma_1 (x) $ is an element-wise nonlinearity (e.g. rectified linear-unit) and $\sigma_2 (x)$ is a normalisation function. For example, one can apply sigmoid to restrict the range to $[0, 1]$, or one can apply softmax operation $\alpha_i^l = e^{q_{att,i}^l}/\sum_i e^{q_{att,i}^l}$ such that the attention map sums to 1. AG is therefore characterised by a set of parameters $\bm{\Theta}_{att}$ containing: linear transformations $\bm{W}_x \in \R^{F_l \times F_{int}}$, $\bm{W}_g \in  \R^{F_g \times F_{int}}$, $\bm{\psi} \in \R^{F_{int} \times 1}$ and bias terms $b_{\psi} \in \R$ , $\bm{b}_{xg} \in \R^{F_{int}}$. The linear transformations are computed using channel-wise $1\times1\times1$ convolutions. 

We note that AG parameters can be trained with the standard back-propagation updates without a need for sampling based optimisation methods as used in hard-attention \citep{mnih2014recurrent}. While AG does not require auxiliary loss function to optimise, we found that using deep-supervision \citep{lee2015deeply} encourages the intermediate feature-maps to be semantically discriminative at each image scale. This ensures that attention units, at different scales, have an ability to influence the responses to a large range of image foreground content. \bluereview{We therefore prevent dense predictions from being reconstructed from small subsets of gated feature-maps.}

\subsubsection{Multi-dimensional Attention} In case of where multiple semantic classes are present in the image, one can learn multi-dimensional attention coefficients. This is inspired by the approach of \citet{shen2017disan}, where multi-dimensional attention coefficients are used
to learn sentence embeddings. Thus, each AG learns to focus on a subset of
target structures. In case of multi-dimensional AGs, each $\alpha^l$ corresponds to a vector and produce $\hat{\bm{x}}^l = [\alpha^{l}_{(1)} \odot \bm{x}^l, \dots, \alpha^{l}_{(m)} \odot \bm{x}^l]$
\bluereview{where $\alpha^{l}_{(k)}$ is $k$-th sub AG and $\odot$ is element-wise multiplication operation.} In each sub-AG, complementary information is extracted and fused to define the output of skip connection. 

\subsubsection{Gating Signal and Grid Attention} As the gating signal $\mb{g}$ must encode global information from large spatial context, it is usually obtained from the coarsest scale activation map. For example in classification, one could use the activation map just before the final softmax layer. In the context of medical imaging, however, since most objects of interest are highly localised, flattening may have the disadvantage of losing important spatial context. In fact, in many cases a few max-pooling operations are sufficient to infer the global context without explicitly using the global pooling. Therefore, we propose a \emph{grid attention} mechanism. The idea is to use the coarse scale feature map before any flattening is done. For example, given an input tensor size of $F_l \times H_x \times W_x$, after $r$ max pooling operations, the tensor size is reduced to $F_g \times H_g \times W_g = F_g \times H_x / (2^r) \times W_y / (2^r)$. To generate the attention map, we can either downsample or upsample the coarse grid to match the spatial resolution of $\bm{x}^l$. In this way, the attention mechanism has more flexibility in terms of what to focus on a regional basis. For upsampling, we chose to use bilinear upsampling. Note that the upsampling can be replaced by a learnable weight, however, we did not opt for this for the sake of simplicity. For segmentation, one can directly use the coarsest activation map as the gating signal. 

\begin{figure}[!t]
	\centering
	\includegraphics[width=0.95\textwidth]{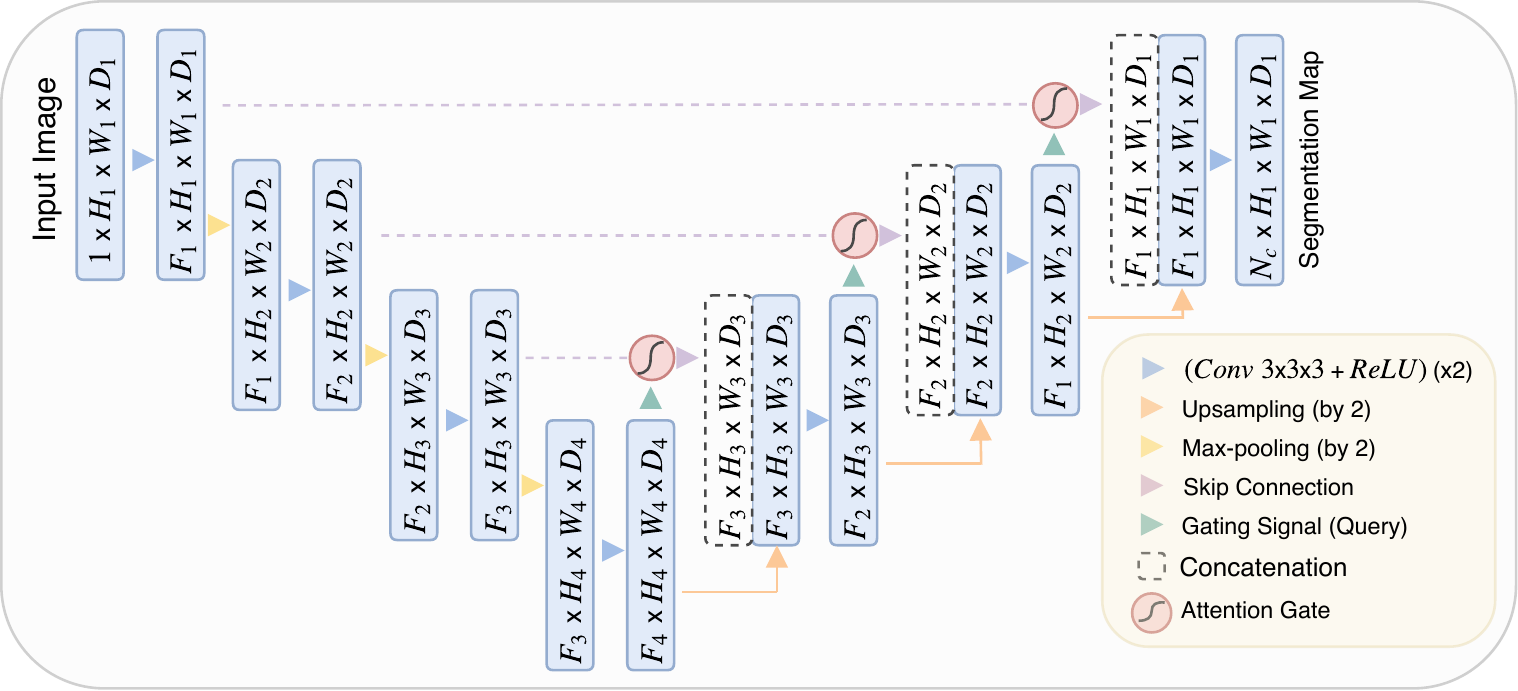}
	\caption{A block diagram of the proposed Attention U-Net segmentation model. Input image is progressively filtered and downsampled by factor of $2$ at each scale in the encoding part of the network (e.g. $H_4=H_1/8$). $N_c$ denotes the number of classes. Attention gates (AGs) filter the features propagated through the skip connections. Schematic of the AGs is shown in Figure \ref{fig:attentionblock}. Feature selectivity in AGs is achieved by use of contextual information (gating) extracted in coarser scales.}
	\label{fig:modelschematic_unet}
\end{figure}

\subsubsection{Backward Pass through Attention Gates} Information extracted from coarse scale is used in gating to disambiguate irrelevant and noisy responses \bluereview{in input feature-maps. For instance, in the U-Net architecture, gating is performed on skip connections right before the concatenation to merge only relevant activations.} Additionally, AGs filter the neuron activations during the forward pass as well as during the backward pass. Gradients originating from background regions are down weighted during the backward pass. This allows model parameters in shallower layers to be updated mostly based on spatial regions that are relevant to a given task. The update rule for convolution parameters in layer $l-1$ can be formulated as follows: \vspace{3.0 mm}
\begin{eqnarray}
\frac{\partial (\hat{x}_i^l)}{\partial \, (\Phi^{l-1})}= \frac{\partial \left(\alpha_i^l \, f(x_i^{l-1}; \Phi^{l-1})\right)}{\partial \, (\Phi^{l-1})} 
= \alpha_i^l \, \frac{\partial(f(x_i^{l-1}; \Phi^{l-1}))}{\partial \, (\Phi^{l-1})} + \frac{\partial (\alpha_i^l)}{\partial \, (\Phi^{l-1})} \, x_i^l
\end{eqnarray}
where the first gradient term on the right-hand side is scaled with $\alpha_i^l$. 

\subsection{Attention Gates for Segmentation}

\blue{In this work, we build our attention-gated segmentation model on top of a standard 3D U-Net architecture}. U-Nets are commonly used for image segmentation tasks because of their good performance and efficient use of GPU memory. The latter advantage is mainly linked to extraction of image features at multiple image scales. Coarse feature-maps capture contextual information and highlight the category and location of foreground objects. Feature-maps extracted at multiple scales are later merged through skip connections to combine coarse- and fine-level dense predictions as shown in Figure \ref{fig:modelschematic_unet}. The proposed AGs are incorporated into the standard U-Net architecture to highlight salient features that are passed through the skip connections. For AGs, we chose sigmoid activation function for normalisation: $\sigma_2(x) = \frac{1}{1+exp(-x)}$. While in image captioning \citep{anderson2017bottom} and classification \citep{jetley2018learn} tasks, the softmax activation function is used to normalise the attention coefficients $\sigma_2$, however, sequential use of softmax yields sparser activations at the output. For dense prediction task, we empirically observed that sigmoid resulted in better training convergence for the AG parameters. 


\subsection{Attention Gates for Classification} 

\begin{figure}[!t]
	\centering
	\includegraphics[width=\textwidth]{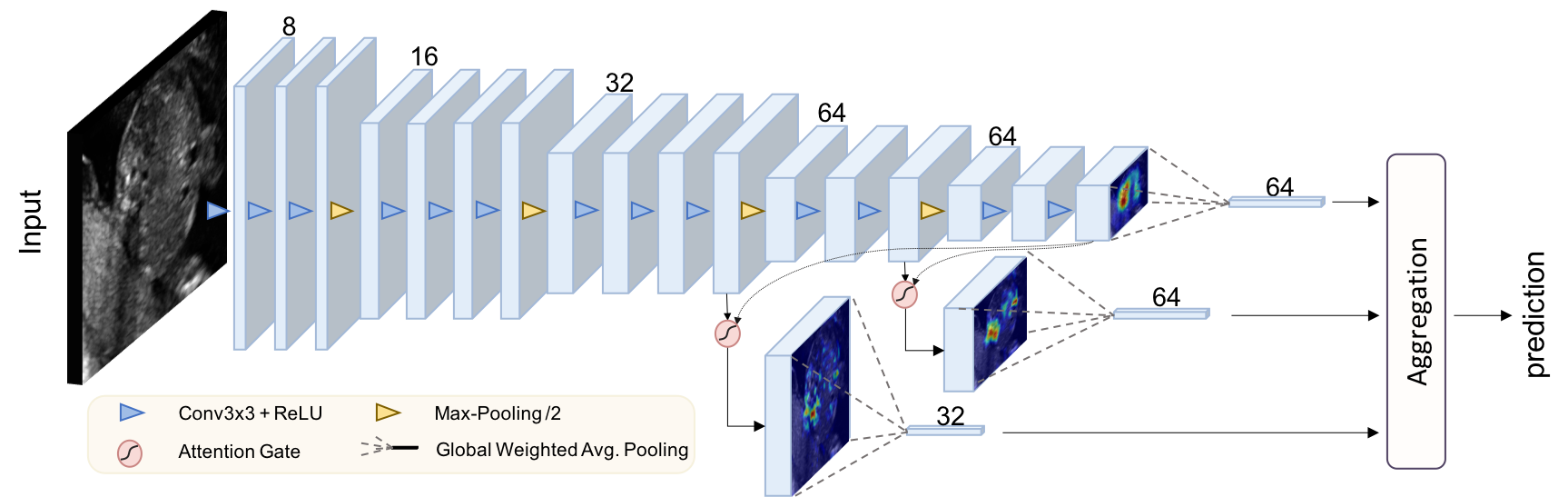}
	\caption{The schematics of the proposed attention-gated classification model, \emph{AG-Sononet}. The proposed attention units are incorporated in layer 11 and layer 14. The attention maps are summed along the spatial axes, resulting in vectors with $F_{l_i}$ features. The vectors are combined using fully connected layers at aggregation stage to yield final predictions.}
	\label{fig:modelschematic}
\end{figure}

For attention-gated classification model, we chose \emph{Sononet} \citep{baumgartner2016real} to be our base architecture, which is a variant of VGG network \citep{simonyan2014very}. The difference is that Sononet can be decoupled into feature extraction module and adaptation module. In the adaptation module, the number of channels are first reduced to the number of target classes $C$. Subsequently, the spatial information is flattened via channel-wise global average pooling. Finally, a softmax operation is applied to the resulting vector and the entry with maximum activation is selected as the prediction. As the network is constrained to classify based on the reduced vector, the network is forced to extract the most salient features for each class.

The proposed attention mechanism is incorporated in the Sononet architecture to better exploit local information. In the modified architecture, termed Attention-Gated Sononet (\emph{AG-Sononet}), we remove the adaptation module. The final layer of the feature extraction module is used as gridded global feature map $\mb{g}$. We apply the proposed attention mechanism to layer 11 and 14 just before pooling, We empirically found that attention gates were less effective if applied to the earliest layer. \bluereview{We speculate that this is because first few layers only represent low-level features, which is not discriminative yet to be attended. The proposed architecture is shown in Figure \ref{fig:modelschematic}}. After the attention coefficients $\{\alpha_i^l\}_{i=1}^n$ are obtained, the weighted average over the spatial axes is computed, yielding a vector of length $F_l$ at scale $l$: $\tilde{\bm{x}}^l = \sum_{i=1}^n \alpha_i^l x_i^l$. In addition, we also perform the global average pooling on the coarsest scale representation. The prediction is given by fitting a fully connected layer on the concatenated feature vector $\{ \tilde{\bm{x}}^{l_1}, \tilde{\bm{x}}^{l_2}, \tilde{\bm{x}}^{l_3} \}$ (e.g. $l_1=11, l_2=14, l_3=17$). We note that for AG-sononet, we normalised the attention coefficients as $\alpha^l_i = (\alpha^l_i - \alpha^l_{min} / \sum_j (\alpha_j^l- \alpha_{min}^l))$, where $\alpha_{min}^l = \min_j \alpha_{j}^l $, as we realised that softmax output was often too sparse, making the prediction more challenging. 

\bluereview{Given the attended feature vectors at different scales, we highlight that the aggregation strategy is flexible and that it can be adjusted depending on the target problem. We empirically observed that a combination of deep-supervision \citep{lee2015deeply} for each scale followed by fine-tuning using a new FC layer fitted on the concatenated vector gave the best performance.}

\redreview{The simplest is to just fit a fully connected layer on the concatenated vector as mentioned above. However, we noticed that sometimes the network abandons the fine-scale attention mechanisms as it is non-trivial to train. An alternative approach is to fit a separate fully connected (FC) layer at each scale and make separate predictions. The final prediction is then given by either weighted mean or max operations. One can also use deep-supervision \mbox{\citep{lee2015deeply}} to force each scale to learn a useful prediction as well as when combined. We empirically observed that first training the network at each scale, then fine-tuning using a new FC layer fitted on the concatenated vector worked the best. In the experimentation, we considered the following variations: }



\section{Experiments and Results}

The proposed AG model is modular and independent of application type; as such it can be easily adapted for pixel and image level classification tasks. To demonstrate its applicability to image classification and segmentation, we evaluate the proposed attention based FCN models on challenging abdominal CT multi-label segmentation and 2D ultrasound image plane classification problems. In particular, pancreas boundary delineation is a difficult task due to shape-variability and poor tissue contrast, similarly image quality and subject variability introduce challenges in 2D-US image classification. Our models are compared against the standard 3D U-Net and Sononet in terms of model prediction performance, model capacity, computation time, and memory requirements. 

\subsection{Evaluation Datasets}
In this section, we present the image datasets used in classification and segmentation experiments. 

\subsubsection{3D-CT Abdominal Image Datasets} 
For the experiments, two different CT abdominal datasets are used: (I) 150 abdominal 3D CT scans acquired from patients diagnosed with gastric cancer ($CT$-$150$). In all images, the pancreas, liver, and spleen boundaries were semi-automatically delineated by three trained researchers and manually verified by a clinician. The same dataset is used by \citet{roth2017hierarchical} to benchmark the U-Net model in pancreas segmentation. (II) The second dataset\footnote{\url{https://wiki.cancerimagingarchive.net/display/Public/Pancreas-CT}} ($CT$-$82$) consists of 82 contrast enhanced 3D CT scans with pancreas manual annotations performed slice-by-slice. This dataset (NIH-TCIA) \citep{tciapancreas} is publicly available and commonly used to benchmark CT pancreas segmentation frameworks. The images from both datasets are downsampled to isotropic $2.00$ mm resolution due to the large image size and hardware memory limitations.  

\subsubsection{2D Fetal Ultrasound Image Dataset} 
Our dataset consisted of 2694 2D ultrasound examinations of volunteers with gestational ages between 18 and 22 weeks. The dataset contains 13 types of standard scan planes and background, complying the standard specified in the UK National Health Service (NHS) fetal anomaly screening programme (FASP) handbook \citep{FASP}. The standard scan planes are: Brain (Cb.), Brain (Tv.), Profile, Lips, Abdominal, Kidneys, Femur, Spine (Cor.), Spine (Sag.), 4CH, 3VV, RVOT, LVOT. The dataset further includes large portions of frames which contains anatomies that are not part of the scan plane, labelled as ``background''. The details of the image acquisition protocol as well as how scan plane labels are obtained can be found in \citep{baumgartner2016real}. The data was cropped to central $208 \times 272$ to prevent the network from learning the surrounding annotations shown in the ultrasound scan screen.


\subsection{Model Training and Implementation Details}

The datasets used in this manuscript contain large class imbalance issue that needs to be addressed. For ultrasound dataset, due to the nature of screening process, the background label dominates the dataset. To address this, we used a weighted sampling strategy, where we matched the probability of sampling one of the foreground labels to the probability of sampling a background label. 
For the segmentation models, the class imbalance problem is tackled \blue{using the Sorensen-Dice loss \citep{drozdzal2016importance, milletari2016v} defined over all semantic classes}. Dice loss is experimentally shown to be less sensitive to class imbalance in segmentation tasks. 

For both tasks, \blue{batch-normalisation, deep-supervision \citep{lee2015deeply}, and standard data-augmentation techniques (affine transformations, axial flips, random crops) are used in training attention and baseline networks.} Intensity values are linearly scaled to obtain a normal distribution $N(0,1)$. For classification models, we empirically found that optimising with Stochastic Gradient Descent with Nesterov momentum ($\rho=0.9$) worked the best. The initial learning rate was set to 0.1, which was subsequently reduced by a factor of 0.1 for every 100 epoch. We also used a warm-start learning rate of 0.01 for the first 5 epochs. For segmentation models, we used Adam with $\alpha=10^{-4}, \beta_1=0.9, \beta_2=0.999$. The batch size for the Sononet models was set to 64. However, for the 3D-CT segmentation models, gradient updates are computed using small batch sizes of $2$ to $4$ samples. For larger segmentation networks, gradient averaging is used over multiple forward and backward passes. This is mainly because we propose a 3D-model to capture sufficient semantic context in contrast to the state-of-the-art CNN segmentation frameworks \citep{cai2017improving, roth2018media}. Gating parameters are initialised so that attention gates pass through feature vectors at all spatial locations. Moreover, we do not require multiple training stages as in hard-attention based approaches therefore simplifying the training procedure.

\begin{table}[t]\footnotesize
	\blue{
	\parbox{\textwidth}{
    	\centering
		\captionof{table}{Multi-class CT abdominal segmentation results obtained on the $CT$-$150$ dataset: The results are reported in terms of Dice score (DSC) and mesh surface to surface distances (S2S). These distances are reported only for the pancreas segmentations. The proposed Attention U-Net model is benchmarked against the standard U-Net model for different training and testing splits. Inference time (forward pass) of the models are computed for input tensor of size $160\times160\times96$. Statistically significant results are highlighted in bold font.}
		\label{tab:results_attention_1}
		\vspace{2 mm}
        \footnotesize
		\begin{tabular}{@{\extracolsep{1pt}}lcccc@{}} 
			Method & U-Net & Att U-Net & U-Net & Att U-Net \\ \midrule
            Train/Test Split &  120/30 & 120/30 & 30/120 & 30/120 \\ \midrule 
			
			Pancreas DSC       &   0.814$\pm$0.116   & \textbf{0.840$\pm$0.087} &  0.741$\pm$0.137 & \textbf{0.767$\pm$0.132} \\
			Pancreas Precision &   0.848$\pm$0.110   & 0.849$\pm$0.098          &  0.789$\pm$0.176 & \textbf{0.794$\pm$0.150} \\
			Pancreas Recall    &   0.806$\pm$0.126   & \textbf{0.841$\pm$0.092} &  0.743$\pm$0.179 & \textbf{0.762$\pm$0.145} \\
			Pancreas S2S Dist (mm) & 2.358$\pm$1.464 & \textbf{1.920$\pm$1.284} &  3.765$\pm$3.452 & 3.507$\pm$3.814 \\ \midrule
			Spleen DSC     &  0.962$\pm$0.013 & 0.965$\pm$0.013 & 0.935$\pm$0.095 & \textbf{0.943$\pm$0.092} \\
			Kidney DSC     &  0.963$\pm$0.013 & 0.964$\pm$0.016 & 0.951$\pm$0.019 & 0.954$\pm$0.021 \\
			Number of Params    &  5.88 M          & 6.40 M          & 5.88 M  & 6.40 M \\
			Inference Time &  0.167 s         & 0.179 s         & 0.167 s & 0.179 s\\ 			\midrule
		\end{tabular}
	}}
\end{table}

\subsubsection{Implementation Details:} 
The architecture for AG-sononet is shown in Fig. \ref{fig:modelschematic}. The parameters for AG-Sononet was initialised using a partially trained Sononet. We compare our models with different capacities, with the initial number of features 8, 16 and 32. For U-net and Attention U-net, the initial number of features is set to $F_1=8$, which is doubled after every max-pooling operation. Our implementation using \blue{PyTorch \citep{paszke2017automatic} is publicly} available\footnote{\url{https://github.com/ozan-oktay/Attention-Gated-Networks}}.

\begin{figure}[!t]
	\centering
	\includegraphics[width=\textwidth]{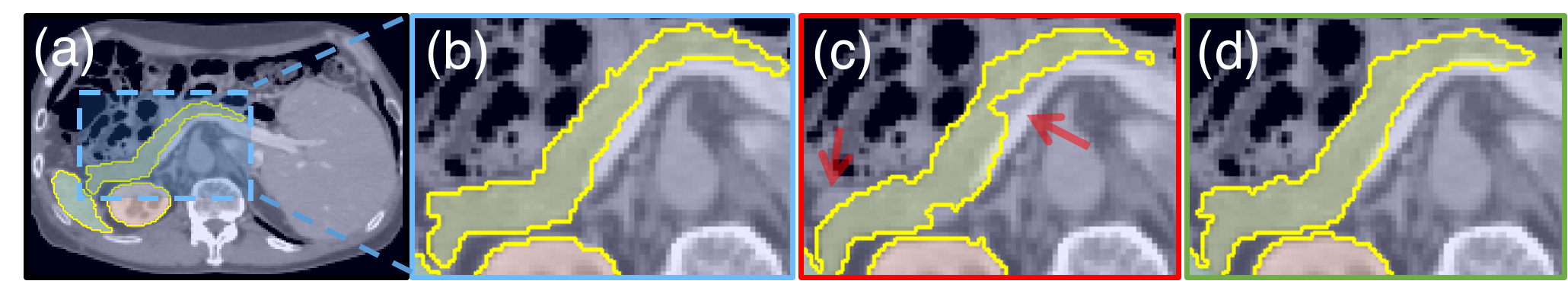}
	\caption{\linespread{1.2}\small (a-b) The ground-truth pancreas segmentation, (c) U-Net and (d) Attention U-Net. The missed dense predictions by U-Net are highlighted with red arrows.}
	\label{fig:qualitative_comparison}
\end{figure}

\subsection{3D-CT Abdominal Image Segmentation Results}

The proposed Attention U-Net model is benchmarked against the standard U-Net \citep{ronneberger2015u} on multi-class abdominal CT segmentation. We use $CT$-$150$ dataset for both training ($120$) and testing ($30$). The corresponding Dice scores (DSC) and surface distances (S2S) are given in Table \ref{tab:results_attention_1}. The results on pancreas predictions demonstrate that attention gates (AGs) increase recall values ($p = .005$) by improving the model's expression power as it relies on AGs to localise foreground pixels. The difference between predictions obtained with these two models are qualitatively compared in Figure \ref{fig:qualitative_comparison}. In the second experiment, the same models are trained with fewer training images ($30$) to show that the performance improvement is consistent and significant for different sizes of training data ($p = .01$). For both approaches, we observe a performance drop on spleen DSC as the training size is reduced. The drop is less significant with the proposed framework. For kidney segmentation, the models achieve similar accuracy since the tissue contrast is higher.

\begin{table}[t]\footnotesize
	\parbox{\textwidth}{
    	\footnotesize
		\centering
		\captionof{table}{Segmentation experiments on $CT$-$150$ dataset are repeated with higher capacity U-Net models to demonstrate the efficiency of the attention models with similar or less network capacity. The additional filters in the U-Net model are distributed uniformly across all the layers. Segmentation results for the pancreas are reported in terms of dice score, precision, recall, surface distances. \blue{The models are trained with the same train/test data splits (120/30)}.}
		\label{tab:results_attention_3}
		\vspace{1 mm}
		\begin{tabular}{@{\extracolsep{1pt}}lcccccc@{}} 
			Method  &\# of Pars & DSC & Precision & Recall & S2S Dist (mm) & Run Time   \\ \midrule
			U-Net   & 6.44 M    & .821$\pm$.119 & .849$\pm$.111 & .814$\pm$.125 & 2.383$\pm$1.918 & .191 s\\
			U-Net   & 10.40 M   & .825$\pm$.104 & .861$\pm$.082 & .807$\pm$.121 & 2.202$\pm$1.144 &  .222 s\\
			\midrule
		\end{tabular}
	}
\end{table}
\begin{table}[t]\footnotesize
	\parbox{\textwidth}{
    	\footnotesize
		\centering
		\captionof{table}{Pancreas segmentation results obtained on the TCIA Pancreas-CT Dataset \citep{tciapancreas}. The dataset contains in total 82 scans which are split into training (61) and testing (21) sets. The corresponding results are obtained before (BFT) and after fine tuning (AFT) and also training the models from scratch (SCR). Statistically significant results are highlighted in bold font.} 
		\label{tab:results_attention_4}
		\vspace{1 mm}
        }	
        \parbox{\textwidth}{
        \footnotesize
		\begin{tabular}{@{\extracolsep{1pt}}llcccccccc@{}} 
        	\footnotesize
			& Method  & Dice Score & Precision & Recall & S2S Dist (mm)  \\ \midrule
			\parbox[t]{2mm}{\multirow{2}{*}{\rotatebox[origin=c]{90}{\underline{BFT}}}}
			& U-Net           & 0.690$\pm$0.132 & 0.680$\pm$0.109 & 0.733$\pm$0.190 & 6.389$\pm$3.900\\
			& Attention U-Net & \textbf{0.712$\pm$0.110} & 0.693$\pm$0.115 & \textbf{0.751$\pm$0.149} & \textbf{5.251$\pm$2.551}\\
			\midrule
			\parbox[t]{2mm}{\multirow{2}{*}{\rotatebox[origin=c]{90}{\underline{AFT}}}}
			& U-Net           & 0.820$\pm$0.043 & 0.824$\pm$0.070 & 0.828$\pm$0.064 & 2.464$\pm$0.529\\
			& Attention U-Net & \textbf{0.831$\pm$0.038} & 0.825$\pm$0.073 & \textbf{0.840$\pm$0.053} &  \textbf{2.305$\pm$0.568}\\ 
			\midrule
			\parbox[t]{2mm}{\multirow{2}{*}{\rotatebox[origin=c]{90}{\underline{SCR}}}}
			& U-Net           & 0.815$\pm$0.068 & 0.815$\pm$0.105 & 0.826$\pm$0.062 & 2.576$\pm$1.180 \\
			& Attention U-Net & 0.821$\pm$0.057 & 0.815$\pm$0.093 & \textbf{0.835$\pm$0.057} & \textbf{2.333$\pm$0.856} \\
			\midrule
		\end{tabular}
	}
\end{table}

In Table \ref{tab:results_attention_1}, we also report the number of trainable parameters for both models. We observe that by adding 8\% extra capacity to the standard U-Net, the performance can be improved by 2-3\% in terms of DSC. For a fair comparison, we also train higher capacity U-Net models and compare against the proposed model with smaller network size. The results shown in Table \ref{tab:results_attention_3} demonstrate that the addition of AGs contributes more than simply increasing model capacity (uniformly) across all layers of the network ($p=.007$). Therefore, additional capacity should be used for AGs to localise tissues, in cases when AGs are used to reduce the redundancy of training multiple, individual models.

\subsubsection{Comparison to State-of-the-Art CT Abdominal Segmentation Frameworks} 

\begin{table}\footnotesize
	\parbox{\textwidth}{
		\centering
		\captionof{table}{State-of-the-art CT pancreas segmentation methods that are based on single and multiple CNN models. The listed segmentation frameworks are evaluated on the same public benchmark ($CT$-$82$) using different number of training and testing images. Similarly, the FCN approach proposed in \citep{roth2017hierarchical} is benchmarked on $CT$-$150$ although it is trained on an external dataset (Ext).}
		\label{tab:state_of_the_art_methods} }
     \resizebox{\linewidth}{!}{
		\vspace{1 mm}
		\begin{tabular}{@{\extracolsep{1pt}}lcccc@{}} 
			Method  & Dataset & Pancreas DSC & Train/Test & \# Folds \\ \midrule
			
			Hierarchical 3D FCN \citep{roth2017hierarchical} & $CT$-$150$ & $82.2\pm10.2$ & Ext/$150$ & -\\
			
			Dense-Dilated FCN \citep{gibson2017towards} & $CT$-$82$ \& Synapse\footnote{\url{https://www.synapse.org/Synapse:syn3193805}} & $66.0\pm10.0$ & $63$/$9$ & 5-CV\\
			
			2D U-Net \citep{heinrich2018ternarynet} & $CT$-$82$ & $75.7\pm9.0$ & 66/16 & 5-CV\\ 
			
			HN 2D FCN Stage-1\citep{roth2018media} & $CT$-$82$ & $76.8 \pm 11.1$ & 62/20& 4-CV\\
			
			HN 2D FCN Stage-2\citep{roth2018media} & $CT$-$82$ & $81.2 \pm 7.3$ & 62/20& 4-CV\\
			
			2D FCN \citep{cai2017improving} & $CT$-$82$ & $80.3\pm9.0$ & 62/20 & 4-CV\\
			
			2D FCN + RNN \citep{cai2017improving} & $CT$-$82$ & $82.3 \pm 6.7$ & 62/20 & 4-CV\\
			
			Single Model 2D FCN \citep{zhou2017fixed} & $CT$-$82$ & $75.7 \pm 10.5$ & 62/20 & 4-CV\\
			
			Multi-Model 2D FCN \citep{zhou2017fixed} & $CT$-$82$ & $82.2 \pm 5.7$ & 62/20 & 4-CV\\
			
			\midrule
		\end{tabular}
	}
\end{table}

The proposed architecture is evaluated on the public TCIA CT Pancreas benchmark to compare its performance with state-of-the-art methods. Initially, the models trained on $CT$-$150$ dataset are directly applied to $CT$-$82$ dataset to observe the applicability of the two models on different datasets. The corresponding results (BFT) are given in Table \ref{tab:results_attention_4}. U-Net model outperforms traditional atlas techniques \citep{wolz2013automated} although it was trained on a disjoint dataset. Moreover, the attention model performs consistently better in pancreas segmentation across different datasets. These models are later fine-tuned (AFT) on a subset of TCIA dataset ($61$ train, $21$ test). The output nodes corresponding to spleen and kidney are excluded from the output softmax computation, and the gradient updates are computed only for the background and pancreas labels. The results in Table \ref{tab:results_attention_4} and \ref{tab:state_of_the_art_methods} show improved performance compared to concatenated multi-model CNN approaches \citep{cai2017improving, roth2018media, zhou2017fixed} due to additional training data and richer semantic information (e.g. spleen labels). \blue{Additionally, we trained the two models from scratch (SCR) with $61$ training images randomly selected from the $CT$-$82$ dataset.} Similar to the results on $CT$-$150$ dataset, AGs improve the segmentation accuracy and lower the surface distances ($p = .03$) due to increased recall rate of pancreas pixels ($p = .09$). 

Results from state-of-the-art CT pancreas segmentation models are summarised in Table \ref{tab:state_of_the_art_methods} for comparison purposes. Since the models are trained on the same training dataset, this comparison gives an insight on how the attention model compares to the relevant literature. It is important to note that, post-processing (e.g. using conditional random field) is not utilised in our framework as the experiments mainly focus on quantification of performance improvement brought by AGs in an isolated setting. Similarly, residual and dense connections can be used as in \citep{gibson2017towards} in conjunction with AGs to improve the segmentation results. In that regard, our 3D Attention U-Net model performs similar to the state-of-the-art, despite the input images are downsampled to lower resolution. More importantly, our approach significantly improves the results compared to single-model based segmentation frameworks (see Table \ref{tab:state_of_the_art_methods}). We do not require multiple CNN models to localise and segment object boundaries. \blue{Lastly, we performed $5$-fold cross-validation on the $CT$-$82$ dataset using the Attention U-Net for a better comparison, which achieved $81.48\pm6.23$ DSC for pancreas labels.}

\begin{table}
\bluereview{
\centering
\scriptsize
\caption{Test results for standard scan plane detection. Number of initial filters is denoted by the postfix ``-$n$''. Time taken for forward (Fwd) and backward (Bwd) passes were recorded in milliseconds. }
\begin{tabular}{lcccccc}
Method & Accuracy & F1 & Precision & Recall & Fwd/Bwd ($ms$) & \#Param \\
\midrule
Sononet-8       & 0.969          & 0.899          & 0.878          & 0.922          & 1.36/2.60 & 0.16M 
\\
AG-Sononet-8 & \textbf{0.977} & \textbf{0.922} & \textbf{0.916} & \textbf{0.929}          & 1.92/3.47 & 0.18M 
\\

\hline
Sononet-16       & 0.977          & 0.923          & 0.916          & 0.931          & 1.45/3.92 & 0.65M 
\\
AG-Sononet-16 & \textbf{0.978} & \textbf{0.929} & \textbf{0.924} & \textbf{0.934} & 1.94/5.13 & 0.70M 
\\
\hline
Sononet-32       & 0.979          & 0.931          & 0.924          & \textbf{0.938} & 2.40/6.72 & 2.58M 
\\
AG-Sononet-32 & \textbf{0.980} & \textbf{0.933} & \textbf{0.931} & 0.935          & 2.92/8.68 & 2.79M 
\\
\bottomrule
\end{tabular}
\label{table:result1}}
\end{table}

\begin{table}
\centering
\footnotesize
\caption{Class-wise performance for AG-Sononet-8. In bracket shows the improvement over Sononet-8. Bold highlights the improvement more than 0.02.}
\scalebox{0.9}{
\begin{tabular}{lccc}
                &  Precision             &  Recall                &  F1                     \\ \midrule
Brain (Cb.)     & 0.988	(-0.002)         & 0.982 (-0.002)         & 0.985 (-0.002)          \\
Brain (Tv.)     & 0.980	(0.003)          & 0.990 (0.002)          & 0.985 (0.003)          \\
Profile         & 0.953	\textbf{(0.055)} & 0.962 (0.009)          & 0.958 \textbf{(0.033)} \\
Lips            & 0.976	\textbf{(0.029)} & 0.956 (-0.003)         & 0.966 (0.013)          \\
Abdominal       & 0.963	(0.011)          & 0.961 (0.007)          & 0.962 (0.009)          \\
Kidneys	        & 0.863	\textbf{(0.054)} & 0.902 (0.003)          & 0.882 \textbf{(0.030)} \\
Femur           & 0.975	(0.019)          & 0.976 (-0.005)         & 0.975 (0.007)          \\
Spine (Cor.)    & 0.935	\textbf{(0.049)} & 0.979 (0.000)          & 0.957 \textbf{(0.026)} \\
Spine (Sag.)    & 0.936	\textbf{(0.055)} & 0.979 (-0.012)         & 0.957 \textbf{(0.024)} \\
4CH	            & 0.943	\textbf{(0.035)} & 0.970 (0.007)          & 0.956 \textbf{(0.022)} \\
3VV	            & 0.694	\textbf{(0.050)} & 0.722 (-0.014)         & 0.708 \textbf{(0.021)} \\
RVOT            & 0.691	\textbf{(0.029)} & 0.705 \textbf{(0.044)} & 0.698 \textbf{(0.036)} \\
LVOT            & 0.925	\textbf{(0.022)} & 0.933 \textbf{(0.027)} & 0.929 \textbf{(0.024)} \\
Background	    & 0.995	(-0.001)         & 0.992 (0.007)          & 0.993 (0.003)          \\
\bottomrule
\end{tabular}}
\label{table:result_class}
\end{table}

\subsection{2D Fetal Ultrasound Image Classification Results}

The dataset was split to training ($122,233$), validation ($30,553$) and testing ($38,243$) frames on subject basis. For evaluation, we used \bluereview{macro-averaged} precision, recall, F1, overall accuracy, the number of parameters and execution speed, \bluereview{summarised in Table \ref{table:result1}.} \\ 

In general, AG-Sononet improves the results over Sononet at all capacity levels. In particular, AG-Sononet achieves higher precision. AG-Sononet reduces false positive examples because the gating mechanism suppresses background noise and forces the network to make the prediction based on class-specific features. As the capacity of Sononet is increased, the gap between the methods are tightened, but we note that the performance of AG-Sononet is also close to the one of Sononet with double the capacity. 
In Table \ref{table:result_class}, we show the class-wise F1, precision and recall values for AG-Sononet-8, where the improvement over Sononet is indicated in brackets. We see that the precision increased by around 5\% for kidney, profie and spines. For the most challenging cardiac views, we see on average 3\% improvement for 4CH and 3VV ($p<0.05$). 

\begin{figure}[!t]
	\centering
	\includegraphics[width=1.00\textwidth]{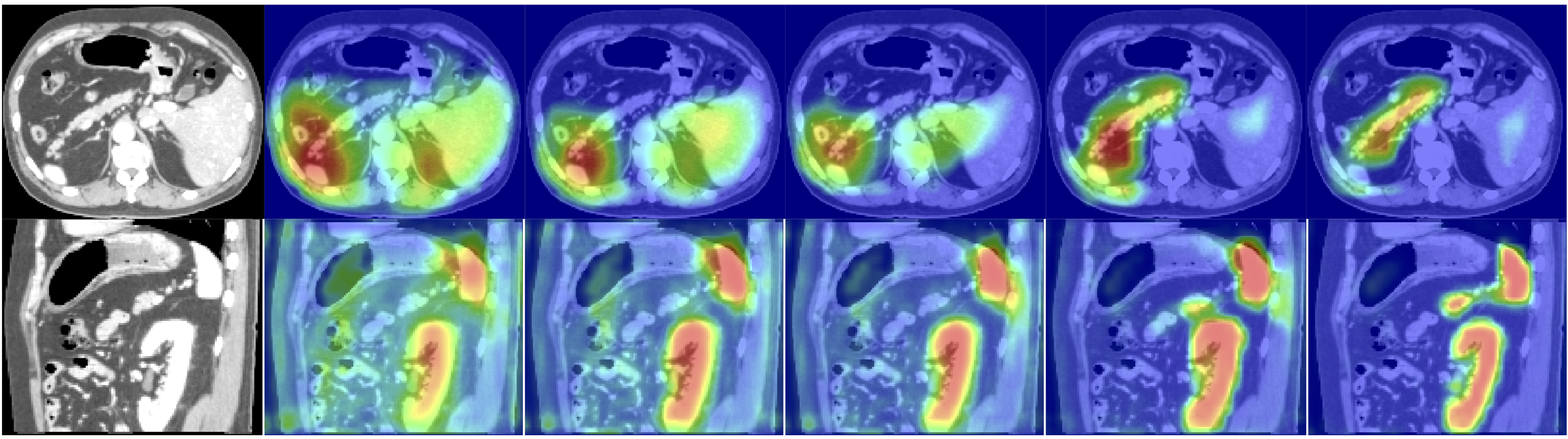}
	\caption{The figure shows the attention coefficients ($\alpha^{l_{s_2}}$, $\alpha^{l_{s_3}}$) across different training epochs ($3$, $6$, $10$, $60$, $150$). The images are extracted from sagittal and axial planes of a 3D abdominal CT scan from the testing dataset. The model gradually learns to focus on the pancreas, kidney, and spleen.}
	\label{fig:attentions_vs_epochs}
\end{figure}

\subsection{Attention Map Analysis} 

The attention coefficients of the proposed U-Net model, which are obtained from 3D-CT test images, are visualised with respect to training epochs (see Figure \ref{fig:attentions_vs_epochs}). We commonly observe that AGs initially have a uniform distribution and pass features at all spatial locations. This is gradually updated and localised towards the targeted organ boundaries. Additionally, at coarser scales AGs provide a rough outline of organs which are gradually refined at finer resolutions. Moreover, by training multiple AGs at each image scale, we observe that each AG learns to focus on a particular subset of organs. 

\subsubsection{Object Localisation using Attention Maps}

\begin{figure}[!t]
	\centering
	\includegraphics[width=1\textwidth]{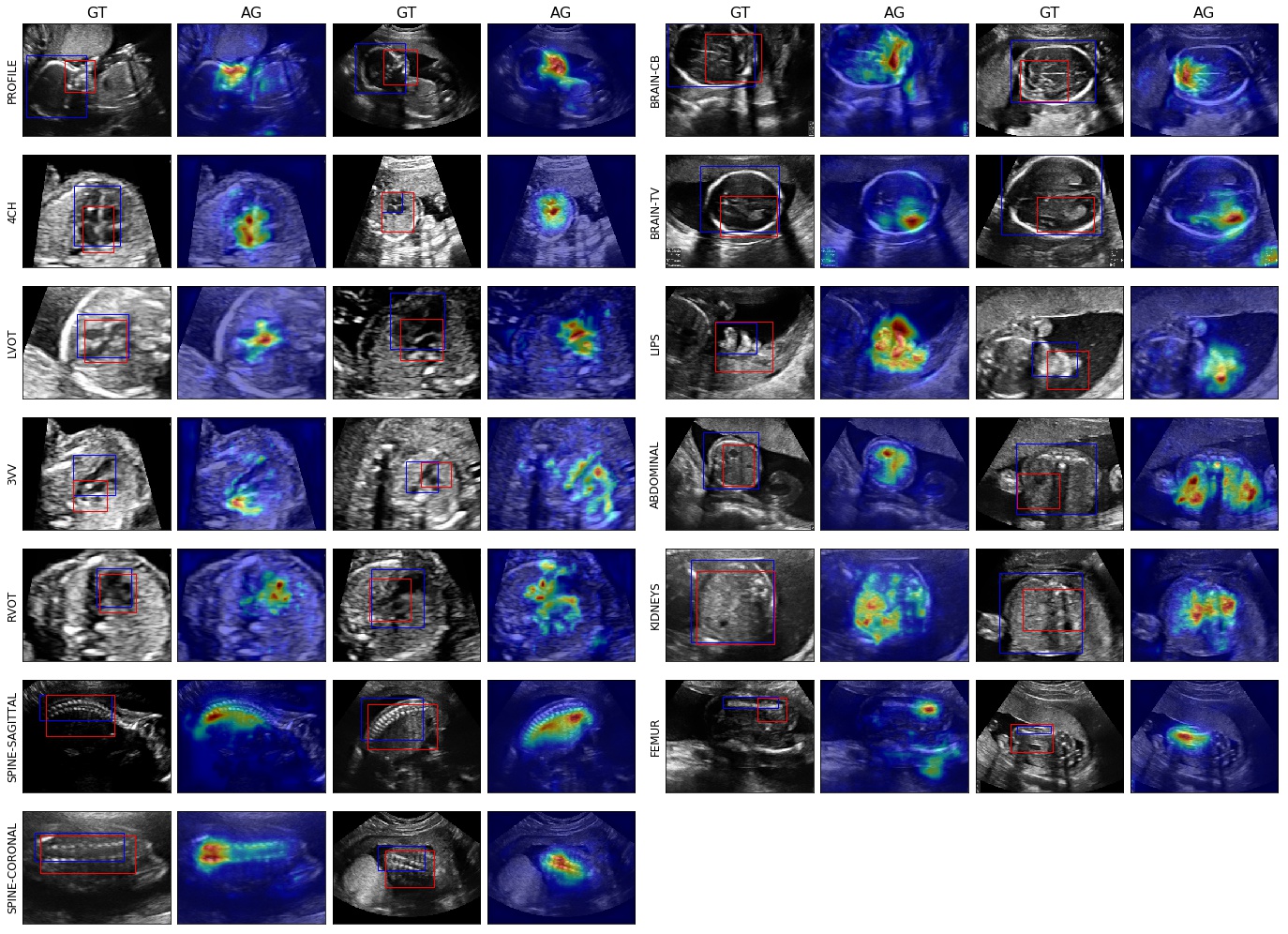}
	\caption{Examples of the obtained attention map and generated bounding boxes (red) from AG-Sononet-FT across different subjects. The ground truth annotation is shown in blue. The detected region highly agrees with the object of interest.}
	\label{fig:attention_ag_sononet_ds_variation}
\end{figure}

\bluereview{With the proposed architecture, the localisation maps can obtained for almost no additional computational cost. In Figure \ref{fig:attention_ag_sononet_ds_variation}, we show the attention maps of AG-Sononet across different subjects, together the red bounding box annotation generated using the attention maps (see Appendix for the heuristics). We see that the network consistently focuses on the object of interest, consistent  with the blue ground truth annotation. We note, however, attention map outlines the discriminant region; in particular, it does not necessarily coincide with the entire object. Nevertheless, as it does not guided backpropagation for localisation (a strategy in  \citep{baumgartner2016real}), attention models are advantageous for the real-time applications.}

\redreview{Finally, in Figure \mbox{\ref{fig:attention_ag_sononet_ds_variation}}, we show the attention maps of AG-Sononet-FT across different subjects, together the bounding box annotation generated using the attention maps (see Appendix for the heuristics). We see that the network consistently focuses on the object of interest, which indicates that the network indeed learnt the most important feature for each class. We note, however, attention map outlines the discriminant region; in particular, it does not necessarily coincide with the entire object. This behaviour makes sense because some part of object will appear in background label (i.e. when the ideal plane is not reached). Qualitatively, however, the bounding boxes well agree with the annotated ground truth. Most crucially, the attention map is obtained for almost no additional computational cost; In comparison, \mbox{\citep{baumgartner2016real}} requires guided backpropagation for localisation, which limits the localisation speed. This highlights the advantage of attention model for the real-time applications.}

\section{Discussion}

In this work, we considered soft-attention mechanism and discussed how to incorporate this idea into segmentation and scan plane detection frameworks to better exploit local structures in CT abdominal and fetal ultrasound images. In particular, we highlighted several aspects: gridded attention mechanisms, a normalisation strategy for the attention map, and aggregation strategies. \blue{We empirically observed and reported that using soft-max as the activation function tends to generate a map that is sparsely activated and is overly sensitive to local intensity changes.} The latter is problematic as in ultrasound imaging, image quality is often low. In the classification setting, We found that dividing the activations by the sum of the activations helped generate attention map with larger contextual support. As demonstrated in the segmentation framework, Sigmoid function is a good alternative as it only normalises the range and allows more information to flow. However, we found that training is non-trivial due to the gradient saturation problem.

We noted that training the attention-mechanism was slightly more complex than the standard network architecture. In particular, we observed that the strategy employed to aggregate the attention maps at different scales affects both the learning of the attention mechanism itself and hence the performance. Having a loss term defined at each scale ensures that the network learns to attend at each scale. We observed that first training the network at each scale separately, followed by fine-tuning was the most stable approach to get the optimal performance. 

There is a vast body of literature in machine learning exploring different gating architectures. For example, highway networks \citep{greff2016highway} make use of residual connections around the gate block to allow better gradient back-propagation and slightly softer attention mechanisms. Although our segmentation experiments with residual connections have not provided any significant performance improvement, future work will focus on this aspect to obtain a better training behaviour.

\blue{Lastly, we note that the presented quantitative comparisons between the Attention 3D-Unet and state-of-the-art 2D cascaded models might not be sufficient enough to draw a final conclusion, as the proposed approach takes advantage of rich contextual information in all spatial dimensions. On the other hand, the 2D models utilise the high resolution information present in axial CT planes without any downsampling. We think that with the advent of improved GPU computation power and memory, larger capacity 3D-CT segmentation models can be trained with larger image grids without the need for image downsampling. In this regard, future research will focus more and more on deploying 3D models, and the performance of Attention U-Net can be further enhanced by utilising fine resolution input batches without any additional heuristics.}

\section{Conclusion}

In this work we proposed a novel and modular attention gate model that can be easily incorporated into existing segmentation and classification architectures. Our approach can eliminate the necessity of applying an external object localisation model by implicitly learning to highlight salient regions in input images. Moreover, in a classification setting, AGs leverage the salient information to perform task adaptive feature pooling operation. 

We applied the proposed attention model to standard scan plane detection during fetal ultrasound screening and showed that it improves overall results, especially precision, with much less parameters. This was done by generating the gating signal to pinpoint local as well as global information that is useful for the classification. Similarly, experimental results on CT segmentation task demonstrate that the proposed AGs are highly beneficial for tissue/organ identification and localisation. This is particularly true for variable small size organs such as the pancreas, and similar behaviour is observed in image classification tasks. 

Additionally, AGs allow one to generate fine-grained attention map that can be exploited for object localisation. We envisage that the proposed soft-attention module could support explainable deep learning, which is a vital research area for medical imaging analysis.  

\bibliography{mybibfile}

\newpage

\section*{Appendix - Weakly Supervised Object Localisation (WSL)}

In \citep{baumgartner2016real}, WSL was performed by exploiting the pixel-level saliency map obtained by guided-backpropagation, followed by ad-hoc procedure to extract bounding boxes. The same heuristics can be applied for the given network, however, owing to the attention map, we can device a much efficient way of performing object localisation. In particular, we generate object location by simply: (1) blur the attention maps, (2) threshold the low activations, (3) perform connected-component analysis, (4) select a component that overlaps at each scale and (5) apply bounding box around the selected components. In this heuristics, backpropagation is not required so it can be executed efficiently. We note, however, attention map outlines salient region used by the network to perform classification; in particular, it does not necessarily agree with the object of interest. This behaviour makes sense because some part of object will appear both in the class as well as background frame until the ideal plane is reached. Therefore, the quantitative result is shown in \ref{table:wsl_result}, however, the result is biased. We however define new metric called \emph{Relative Correctness}, which is defined as 50\% of maximum achievable IOU (due to bias). We see that in this metric, the method achieves very high results, indicating that it can detect relevant features of the object of interest in its proximity. 

\begin{table}[htb]
\centering
\caption{WSL performance for the proposed strategy with AG-Sononet-16. Correctness (Cor.) is defined as $IOU > 0.5$. Relative Correctness (Rel.) is defined as $IOU > 0.5\times \max(IOU_{class})$. }
\scalebox{0.8}{
\begin{tabular}{lccc}
             &  IOU Mean (Std) &  Cor. (\%)  &  Rel. (\%) \\ \midrule
Brain (Cb.)  &   0.69 (0.11) &  0.96 & 0.96  \\
Brain (Tv.)  &   0.68 (0.12) &  0.96 & 0.96  \\
Profile      &   0.31 (0.08) &  0.00 & 0.80  \\
Lips         &   0.42 (0.18) &  0.36 & 0.60  \\
Abdominal    &   0.71 (0.10) &  0.96 & 0.96  \\
Kidneys      &   0.73 (0.13) &  0.92 & 0.98  \\
Femur        &   0.31 (0.11) &  0.02 & 0.58  \\
Spine (Cor.) &   0.53 (0.13) &  0.56 & 0.76  \\
Spine (Sag.) &   0.53 (0.11) &  0.54 & 0.94  \\
4CH          &   0.61 (0.14) &  0.76 & 0.86  \\
3VV          &   0.42 (0.14) &  0.34 & 0.62  \\
RVOT         &   0.56 (0.15) &  0.70 & 0.76  \\
LVOT         &   0.54 (0.15) &  0.62 & 0.80  \\
\bottomrule
\end{tabular}}
\label{table:wsl_result}
\end{table}

\section*{Acknowledgements}
We thank the volunteers, radiographers and experts for providing manually annotated datasets, Wellcome Trust IEH Award [102431], NVIDIA for their GPU donations, and Intel.

\end{document}